\def\year{2020}\relax
\documentclass[letterpaper]{article} 
\usepackage{aaai20}  
\usepackage{times}  
\usepackage{helvet} 
\usepackage{courier}  
\usepackage[hyphens]{url}  
\usepackage{graphicx} 
\usepackage{graphicx}  
\usepackage{amsmath,amssymb,amsthm,bbm,mathtools,bm}
\usepackage{algorithm,algorithmicx,algpseudocode}
\usepackage{subfigure}
\usepackage{color}

\DeclareMathOperator{\diag}{diag}

\DeclareMathOperator{\sigmoid}{sigmoid}
\DeclareMathOperator{\gru}{GRU}

\newcommand{\astar}{\ensuremath{\mathit{A}^{\ast}}}

\newcommand{\pddl}{\textsc{PDDL}}

\title{Online Planner Selection with Graph Neural Networks and Adaptive Scheduling}
\author{
 Tengfei Ma,$^1$
 Patrick Ferber,$^{3,4}$
 Siyu Huo,$^1$
 Jie Chen,$^{1,2}$\thanks{Contact authors}
 Michael Katz$^1$\footnotemark[1]\\
 $^1$IBM Research,
 $^2$MIT-IBM Watson AI Lab,
 $^3$University of Basel,
 $^4$Saarland University\\
 Tengfei.Ma1@ibm.com,
 patrick.ferber@unibas.ch,
 siyu.huo@ibm.com,
 chenjie@us.ibm.com,
 Michael.Katz1@ibm.com
}
 
\begin{document}

\maketitle

\begin{abstract}
Automated planning is one of the foundational areas of AI. Since no
single planner can work well for all tasks and domains,
portfolio-based techniques have become increasingly popular in recent
years. In particular, deep learning emerges as a promising methodology
for online planner selection. Owing to the recent development of
structural graph representations of planning tasks, we propose a graph
neural network (GNN) approach to selecting candidate planners.  GNNs
are advantageous over a straightforward alternative, the convolutional
neural networks, in that they are invariant to node permutations and
that they incorporate node labels for better inference.

Additionally, for cost-optimal planning, we propose a two-stage
adaptive scheduling method to further improve the likelihood that a
given task is solved in time. The scheduler may switch at halftime to
a different planner, conditioned on the observed performance of the
first one. Experimental results validate the effectiveness of the
proposed method against strong baselines, both deep learning and
non-deep learning based.

The code is available at \url{https://github.com/matenure/GNN_planner}.
\end{abstract}

\section{Introduction}

Automated planning is one of the foundational areas of Artificial
Intelligence research~\cite{russell-norvig-1995}. 
Planning is
concerned with devising goal-oriented policies executed by agents in
large-scale state models.  Since planning is intractable in
general~\cite{chapman-aij1987}
and even classical planning is
PSPACE-complete~\cite{bylander-aij1994},
a single algorithm unlikely
works well for all problem domains.  Hence, surging interest exists in
developing portfolio-based
approaches~\cite{seipp-et-al-icaps2012,vallati-miwai2012,howe-et-al-ecp1999,seipp-et-al-aaai2015},
which, for a set of planners, compute an offline schedule or an online
decision regarding which planner to invoke per planning task.  While
offline portfolio approaches focus on finding a single invocation
schedule that is expected to work well across all planning tasks,
online methods {\em learn} to choose the right planner for each given
task.  Most online methods use handcrafted features for
learning~\cite{cenamor-et-al-jair2016}.

Recent advances in deep learning have stimulated increasing interest
in the use of deep neural networks for online portfolio selection,
alleviating the effort of handcrafting features. A deep neural network
may be considered a machinery for learning feature representations of
an input object without the tedious effort of feature engineering. For
example, convolutional neural networks (CNN) take the raw pixels as
input and learn the feature representation of an image through layers
of convolutional transformations and abstractions, which result in a
feature vector that captures the most important characteristics of the
image~\cite{Krizhevsky2012}. A successful example in the context of
planning is
\emph{Delfi}~\cite{katz-et-al-ipc2018,Sievers2019}, which treats a planning task as an image
and applies CNN to predict the probability that a certain planner
solves the task within the time limit. Delfi won the first place in
the Optimal Track of the 2018 International Planning Competition
(IPC).

As planning tasks admit state transition graphs that are often too big
to fit in any conceivable size memory, several other graphs were
developed to encode the structural information. Two prominent examples
are the problem description graph~\cite{pochter-et-al-aaai2011} for
a \emph{grounded} task representation; and the abstract structure
graph~\cite{sievers-et-al-icaps2017wshsdip-a,sievers-et-al-icaps2019} for a \emph{lifted}
representation.  Both graphs are used in classical planning for
computing structural
symmetries~\cite{sievers-et-al-icaps2019,domshlak-et-al-icaps2012}.
The most important use of structural symmetries is search space
pruning, considerably improving the state-of-the-art. The lifted
structural symmetries are also found useful for faster grounding and
mutex generation \cite{roeger-et-al-icaps2018}.

\begin{figure*}[ht]
\centering
\includegraphics[height=1.4in]{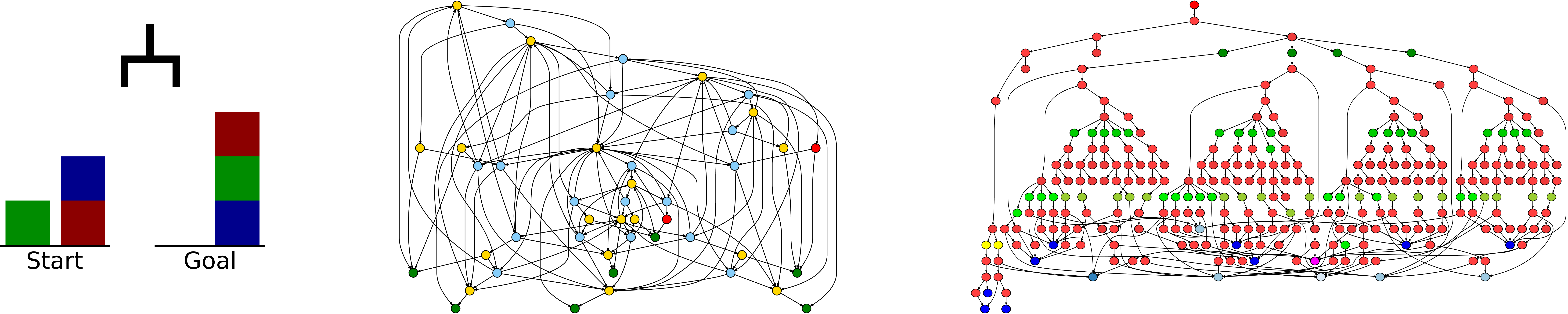}
\caption{An example planning task (left) with its grounded graph representation (middle) and the lifted one (right). The task, {\em blocksworld}, uses a gripper to rearrange a set of blocks from an initial configuration to the goal configuration. The coloring of the graph nodes indicate node labels. For more details, see the section ``Graph Construction.''}
\label{fig:planning}
\end{figure*}

Owing to the development of these structural graphs, we propose a
graph neural network (GNN) approach to learning the feature
representation of a planning task. A proliferation of GNN
architectures emerged
recently~\cite{Bruna2014,Defferrard2016,Li2016,Kipf2017,Hamilton2017,Gilmer2017,Velickovic2018}.
In this work, we explore the use of two representative GNNs---graph
convolutional networks~\cite{Kipf2017} and gated graph neural
networks~\cite{Li2016}. The former is \emph{convolutional}, which
extends convolution filters for image patches to graph
neighborhoods. The latter is \emph{recurrent}, which treats the
representation of a graph node as a dynamical-system state that can be
recurrently updated through neighborhood aggregation. A key difference
between the two is whether network parameters are shared across
layers/steps, similar to that between a CNN and a recurrent neural
network.

GNNs have two advantages over CNNs for graph inputs. First, GNNs
address the limitation of images that are not invariant to node
permutation. Second, GNNs incorporate node and edge attributes that
produce a richer representation than does the image surrogate of the
graph adjacency matrix alone.

With the use of GNNs, we in addition consider the problem of
cost-optimal planning, whose goal is to solve as many tasks as
possible, each given a time limit, with cost-optimal planners. We
propose a two-stage adaptive scheduling approach that enhances the
likelihood of task solving within the time limit, over the usual
approach of using a single planner for the whole time span. The
proposal is based on the observation that if a planner solves a given
task in time, its execution is often rather quick. Hence, we divide
the time interval in two equal halves and determine at the midpoint
whether to change the planner, should it be still running at that
time. Experimental results show that the proposed adaptive scheduling
consistently increases the number of solved tasks, compared with the
use of a single planner.

\section{Planning and Planner Selection}
Planning algorithms generally perform reachability analysis in
large-scale state models, which are implicitly described in a concise
manner via some intuitive declarative
language.
One of the
most popular approaches to classical planning in general and to
cost-optimal planning in particular is state-space heuristic
search. The key to this approach is to automatically derive an
informative heuristic function $h$ from states to scalars, estimating
the cost of the cheapest path from each state to its nearest goal
state. The search algorithms then use these heuristics as search
guides. If $h$ is admissible (that is, it never overestimates the true
cost of reaching a goal state), then search algorithms such as
$\astar$
are guaranteed to provide a
cost-optimal plan.

Over the years, many admissible heuristics were developed to capture
various aspects of planning tasks; see, e.g.,
\cite{edelkamp-ecp2001,helmert-et-al-jacm2014,helmert-domshlak-icaps2009,haslum-et-al-aaai2005}.
Further, search pruning techniques
\cite{wehrle-helmert-icaps2014,shleyfman-et-al-aaai2015}
were developed to reduce the search effort, producing sophisticated
search algorithms
\cite{edelkamp-et-al-aaai2015,gnad-hoffmann-icaps2015}.
All these techniques can be used interchangeably. Moreover, most of
them are highly parameterized, allowing to construct many possible
cost-optimal planners.

Because of the intractability of planning, a single planner unlikely
works well across all possible domains. Some planners excel on certain
tasks, while some on others. However, given a task, it is unclear
whether a particular planner works well on the task without actually
running it.  With a large number of planners, especially in resource
constrained situations, it is infeasible to try all of them until a
good one is found. Hence, it is desirable to predict the performance
of the planners on the task and select the best performing one.

One approach of making such a selection allocates a time budget to
each planner and assigns the same allocation for all tasks,
offline. Prominent examples include Fast Downward Stone Soup
(FDSS)~\cite{helmert-et-al-ipc2011} and Fast Downward
Cedalion~\cite{seipp-et-al-ipc2014b}.

Another approach uses supervised machine learning to predict an
appropriate planner for a task. The predictive model requires for each
task a feature representation, often
handcrafted~\cite{howe-et-al-ecp1999,cenamor-et-al-jair2016},
including for example the number of actions, objects, predicates in
the planning task, and the structure of the task's causal graph. This
approach worked reasonably well in practice for non-optimal planning,
winning the first place in IPC 2014. However, even the updated
version, whose portfolio included top performing planners at IPC 2018
(e.g., the one presented by~\cite{katz-hoffmann-ipc2014}), performed
poorly in this competition, ranked only 12th.

For cost-optimal planning, an online approach is a meta-search in the
space of solution set preserving problem
modifications \cite{fuentetaja-et-al-ipc2018}, aiming at finding task
formulations that a planner would work well on. The resulting
planner, \emph{MSP}, also ranked 12th in IPC 2018.

Yet another cost-optimal planner performing online selection is Delfi,
which treats a planning task as an image and selects a planner from
the portfolio through training a CNN to predict which planner solves
the given task in time. Specifically, a planning task is formulated as
a certain graph, whose adjacency matrix is converted to an image, and
a CNN is used to perform image classification. Delfi won IPC 2018.

Note that graph representations have also been used for probabilistic
planning~\cite{toyer-et-al-aaai2018}, which is applied to a specific
domain. In such a setting, action schemas are shared among the input
tasks, but it is unclear how the approach can be adapted to
domain-independent settings.

\section{Graph Construction}
Two versions of Delfi were submitted to IPC 2018, differing in the way
the planning task is represented. Delfi1 works on the lifted
representation of the task, based on \pddl's abstract structure graph
(ASG)~\cite{sievers-et-al-icaps2019}; whereas Delfi2 works on
the grounded representation, based on the problem description graph
(PDG)~\cite{pochter-et-al-aaai2011}.  Both graphs have additional
features (e.g., node labels), which are ignored when being converted
to an image.

In this work, we reuse the graphs built by Delfi, incorporating
additionally node labels. Figure~\ref{fig:planning} shows a classical
planning example, {\em blocksworld}, with its two graphs. For
illustrative purpose only the three-block version is shown; the
problem is NP-hard.

For the construction of ASGs, planning tasks correspond to abstract
structures, which include actions, axioms, the initial state, and the
goal. Nodes are labeled by their types; e.g., action, axiom,
predicate, and object. Edges encode the inclusion hierarchy of the
abstract structures.

For the construction of PDGs, there are nodes for all task facts,
variables, actions, conditional effects, and axioms. Each node type
has a separate label, further divided by the action cost in the case
of action nodes, and whether the fact is initially true and required
in the goal, in the case of facts. Edges connect facts to their
variables, actions to their conditional effects, conditional effects
to their effect facts, condition facts to their conditional effects,
precondition facts to their actions and axioms, and actions and axioms
to their unconditional effect facts.

\section{Planner Selection with Graph Neural Nets}
Given a portfolio of planners, we model the selection problem as predicting the probability that each planner fails to solve a given task in time. Then, the planner with the lowest probability is selected for execution. Denote by $G$ a task, $\mathcal{G}=\{G\}$ the space of tasks, and $D$ the size of the portfolio. Parameterized by $\theta\in\Theta$, the problem amounts to learning a $D$-variate function $f:\mathcal{G}\times\Theta\to[0,1]^D$ that computes the failure probabilities for all planners in the portfolio.

Let $S=\{(G,y)\}$ be the set of task-label pairs for training, where $y\in\{0,1\}^D$ is the ground-truth labeling vector, whose element $y_j$ denotes the fact whether planner $j$ fails to solve the task in time:
\begin{equation}\label{eqn:y}
y_j=\begin{cases}
0, & \text{if execution time of $j$ does not exceed $T$}, \\
1, & \text{otherwise}.
\end{cases}
\end{equation}
Then, the learning amounts to finding the optimal parameter $\theta$ that minimizes the cross-entropy loss function
\begin{multline*}
L(\theta)=-\sum_{(G,y)\in S,}\sum_{j=1}^D y_j\log f_j(G,\theta)\\
+(1-y_j)\log (1-f_j(G,\theta)).
\end{multline*}

\subsection{Graph Representation Learning}
Since a planning task is formulated as a graph, we write $G=(V,E)$, where $V$ is the node set and $E$ is the edge set. For calculus, the function $f$ requires a vectorial representation $h_G$ of the graph $G$. Deep learning uses deep neural networks to compute this vector, rather than handcrafting. In our work, the design of $f$ consists of three steps:
\begin{enumerate}
\item Parameterize the vectorial representation $h_v$ for all nodes $v\in V$.

\item Form the graph representation as a weighted combination of $h_v$:
\begin{equation}\label{eqn:hG}
h_G=\sum_{v\in V}\alpha_vh_v,
\end{equation}
where $\alpha_v$ denotes the attention weight, scoring in a sense the importance of the contribution of each node to the overall representation of the graph.

\item Parameterize $f$ as a feedforward neural network, taking $h_G$ as input:
\begin{equation}\label{eqn:f}
f(G,\theta)=\sigmoid(W_{\text{logit}}^{\top}h_G).
\end{equation}
The parameter set $\theta$ thus includes the parameter matrix $W_{\text{logit}}$ and all the parameters in $h_v$ and $\alpha_v$.
\end{enumerate}

Graph neural networks differ in the parameterizations of the node representation $h_v$ and possibly the attention weight $\alpha_v$. In this work, we consider two types of GNNs: graph convolutional networks and gated graph neural networks.

\subsection{Graph Convolutional Networks (GCN)}
GCN~\cite{Kipf2017} generalizes the convolution filters for image patches to graph neighborhoods. Whereas an image patch contains a fixed number of pixels, which may be handled by a fixed-size filter, the size of a node neighborhood varies. Hence, the convolution filter for graphs uses a parameter matrix to transform each node representation computed from the past layer, and linearly combines the transformed representations with certain weights based on the graph adjacency matrix.

Specifically, let $t$ be the layer index, orient the node representations $h_v^{(t)}$ as row vectors, and stack them to form the matrix $H^{(t)}$. A layer of GCN is defined as
\[
H^{(t+1)}=\sigma(\widehat{A}H^{(t)}W^{(t)}).
\]
Here, $W^{(t)}$ is the parameter matrix, $\widehat{A}$ is a normalization of the adjacency matrix $A$, and $\sigma$ is an activation function (e.g., ReLU). 
The normalization is defined as
\[
\widehat{A}=\widetilde{D}^{-\frac{1}{2}}\widetilde{A}\widetilde{D}^{-\frac{1}{2}},
\,\, \widetilde{A}=A+I,
\,\, \widetilde{D}=\diag(d_i),
\,\, d_i=\sum_k\widetilde{A}_{ik}.
\]

Using an initial feature matrix $X$ (which, for example, can be defined based on node labels by using one-hot encoding) as the input $H^{(0)}$, a few graph convolutional layers produce a sophisticated representation matrix $H^{(T)}$, whose rows are treated as the final node representations $h_v$. Orient them back as column vectors; then, the attention weights are defined by using a feedforward layer
\begin{equation}\label{eqn:alpha}
\alpha_v=\sigmoid(w_{\text{gate}}^{\top}[h_v^{(T)};h_v^{(0)}]),
\end{equation}
where $w_{\text{gate}}$ is a parameter vector. Hence, the overall parameter set for the model $f$ by using the GCN architecture is
\[
\theta=\{W_{\text{logit}}, w_{\text{gate}}, W^{(0)}, W^{(1)}, \ldots, W^{(T-1)}\}.
\]

\subsection{Gated Graph Neural Networks (GG-NN)}
The architecture of GG-NN~\cite{Li2016} is recurrent rather than convolutional. In this architecture, the node representation is treated as the state of a dynamical system and the gated recurrent unit (GRU) is used to update the state upon a new input message:
\[
h_v^{(t+1)}=\gru(h_v^{(t)},m_v^{(t+1)}).
\]
The message $m_v^{(t+1)}$ is an aggregation of the transformed states for all the neighboring nodes of $v$. Specifically, denote by $\text{in}(v)$ and $\text{out}(v)$ the sets of in-neighbors and out-neighbors of $v$, respectively, and let $W_{\text{in}}$ and $W_{\text{out}}$ be the corresponding parameter matrices shared by all graph nodes and recurrent steps. The message is then defined as
\[
m_v^{(t+1)}=\sum_{u\in \text{in}(v)}W_{\text{in}}^{\top}h_u^{(t)}
+\sum_{u'\in \text{out}(v)}W_{\text{out}}^{\top}h_{u'}^{(t)}.
\]

Similar to GCN, GG-NN may use the initial features for each node as the input $h_v^{(0)}$ and produce $h_v^{(T)}$ as the final node representation $h_v$, through $T$ recurrent steps. Thus, the attention weights $\alpha_v$ may be computed in the same manner as~\eqref{eqn:alpha}. Therefore, the overall parameter set for the model $f$ by using GG-NN is
\[
\theta=\{W_{\text{logit}}, w_{\text{gate}}, W_{\text{in}}, W_{\text{out}}, \text{ and parameters of GRU}\}.
\]

\subsection{Variants}

One variant of the attention weights in~\eqref{eqn:alpha} is that the parameter vector $w_{\text{gate}}$ may not be shared by the planners. In other words, for each planner $j$, a separate parameter $w_{\text{gate},j}$ is used to compute the attention weights $\alpha_{v,j}$ and subsequently the graph representation $h_{G,j}$ and the predictive model $f_j(G,\theta)=\sigmoid(W_{\text{logit},j}^{\top}h_{G,j})$. In this manner, node representations are still shared by different planners, but not the graph representation. Such an approach may be used to increase the capacity of the model $f$, which sometimes works better than using a single $w_{\text{gate}}$.


\section{Adaptive Scheduling}
When the goal is to solve a given task within a time limit $T$ (but not how quickly it is solved), one may try a second planner if she ``senses'' that the selected one unlikely completes in time. Such a scenario may occur when the model $f$ described in the preceding section is insufficiently accurate. Then, we offer a second chance to reevaluate the probability that the currently invoked planner cannot complete within the rest of time allowance, versus the probability that a separate planner fails to solve the task in this time span. If the former probability is lower, we have no choice but to continue the execution of the current planner; otherwise, we switch to the latter one. The intuition comes from the observation that if a planner solves a task in time, often it completes rather quickly. Hence, the remaining time may be sufficient for a second planner, should its failure/success be accurately predicted.

To formalize this idea, we set the time of reevaluation at $T/2$. We learn a separate model $g$ that predicts the probabilities that each planner fails to solve the task before timeout, conditioned on the fact that the current planner $p$ needs more time than $T/2$. We write the function $g:\mathcal{G}\times[D]\times\Theta\to[0,1]^D$, where $[D]$ denotes the set of integers from $1$ to $D$, and parameterize it as
\[
g(G,p,\theta_g)=\sigmoid(W_{\text{logit}}^{\top}h_G+W_{\text{fail}}^{\top}e_p),
\]
where $e_p\in\{0,1\}^D$ is the one-hot vector whose $p$th element is $1$ and $0$ for others.

Compare this model with $f$ in~\eqref{eqn:f}. First, we introduce an additional parameter matrix $W_{\text{fail}}$ to capture the conditional fact. Second, the graph representation $h_G$ reuses that in $f$. In other words, the two models $f$ and $g$ share the same graph representation but differ in the final prediction layer.

\begin{table*}[ht]
\centering
\caption{Summary of data set.}
\label{tab:data.set}
\begin{tabular}{lcc}
\hline
& Grounded & Lifted   \\
\hline
\# Graphs, train+val / test & \multicolumn{2}{c}{2,294 / 145}\\
\# Nodes, min / max / mean / median & 6 / 47,554 / 2,056 / 580 & 51 / 238,909 / 3,001 / 1,294 \\
Node degree, min / max / mean / median & 0.88 / 10.65 / 3.54 / 3.28 & 1.04 / 1.82 / 1.49 / 1.47 \\
\# Node labels & 6 & 15 \\
\hline
\end{tabular}
\end{table*}

\subsection{Training Set}
One must construct a training set $S_g$ for learning the model $g$. One approach is to reuse all the graphs in the training of the model $f$. For every such graph $G$, we pick the planners $p$ whose execution time exceeds $T/2$ and form the pairs $(G,p)$. For each such pair, we further construct the ground-truth labeling vector $z\in\{0,1\}^D$ to form the training set $S_g=\{(G,p,z)\}$.

The construction of the labeling vector follows this rationale: For any planner $j$ different from $p$, because the time allowance is only $T/2$, straightforwardly $z_j=0$ if $j$ solves the task in time less than $T/2$; otherwise, $z_j=1$. On the other hand, when $j$ coincides with $p$, the continued execution of $j$ gives a total time allowance $T$. Hence, $z_j=0$ if $j$ solves the task in time less than $T$ and otherwise $z_j=1$. To summarize,
\[
z_j=
\begin{cases}
0, & \text{if $j=p$ and execution time of $j$ is $\le T$}, \\
1, & \text{if $j=p$ and execution time of $j$ is $>T$}, \\
0, & \text{if $j\ne p$ and execution time of $j$ is $\le T/2$}, \\
1, & \text{if $j\ne p$ and execution time of $j$ is $>T/2$}.
\end{cases}
\]

The size of the training set $S_g$ constructed in this manner may be smaller, but more likely greater, than that of $S$, depending on the performance of the planners on each task. In practice, we find that $|S_g|$ is a few times of $|S|$. Such a size does not incur substantially more expense for training.

With the training set defined, the loss function is
\begin{multline*}
L_g(\theta_g)=-\sum_{(G,p,z)\in S_g,}\sum_{j=1}^D z_j\log g_j(G,p,\theta_g)\\
+(1-z_j)\log (1-g_j(G,p,\theta_g)).
\end{multline*}

\subsection{Two-Stage Scheduling}
We now have two models $f$ and $g$. In test time, we first evaluate $f$ and select the planner $p$ with the lowest predicted probability for execution. If it solves the task before halftime $T/2$, we are done. Otherwise, at halftime, we evaluate $g$ and obtain a planner $j$ with the lowest predicted probability. If $j=p$, we do nothing but to let the planner continue the job. Otherwise, we terminate $p$ and invoke $j$, hoping for a successful execution.

\section{Experiments}

\subsection{Data Set and Portfolio}
To evaluate the effectiveness of the proposals, we prepare a data set composed of historical and the most recent IPC tasks.
Specifically, the historical IPC tasks form the training and validation sets, whereas those of the year 2018 form the test set. A small amount of tasks are ignored, the reason of which is explained in the next paragraph. Summary of the data set is given in Table~\ref{tab:data.set}.

We use the same portfolio as did Delfi~\cite{katz-et-al-ipc2018}, which contains 17 base planners, as it is convenient to compare with the image-based CNN approach. See the cited reference for details of the base planners. The tasks unsolvable by any of these planners within the time limit $T=1800$s are ignored in the construction of the data set. 

We prepare two sets of training/validation splits. The first set reuses the split in Delfi, which conforms to the competition setting where a single model is used for evaluation. On the other hand, to reduce the bias incurred by a single model, for another set we randomly generate 20 splits (with an approximately 9:1 ratio). In ten of them, tasks from the same domain are not separated, whereas in the other ten, they may. We call the former scenario \emph{domain-preserving split} whereas the latter \emph{random split}.

\begin{figure}[ht]
\centering
\includegraphics[width=0.9\linewidth]{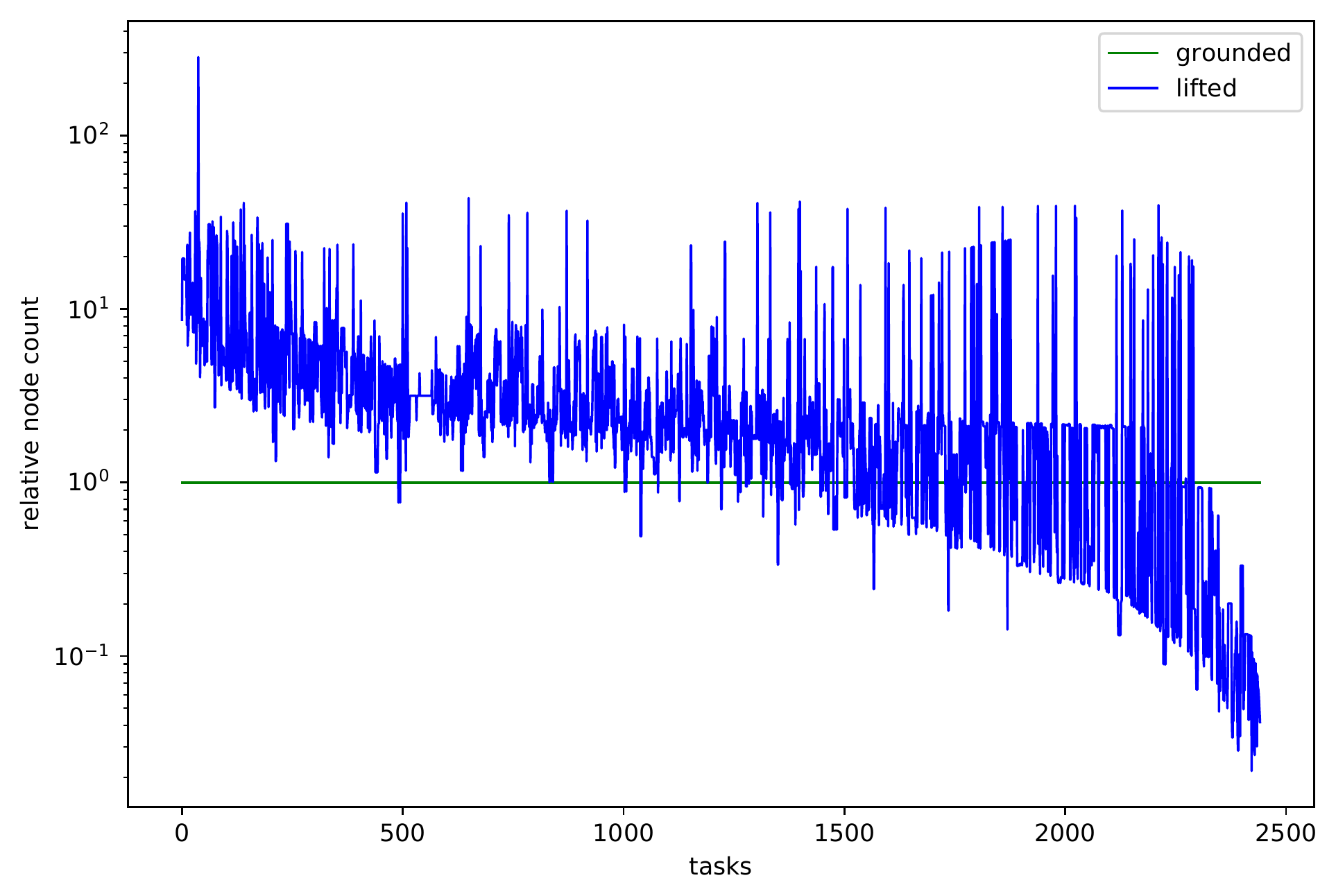}
\caption{Node counts of the lifted graphs relative to that of the grounded ones (tasks sorted in increasing order of grounded size).}
\label{fig:size.comparison}
\end{figure}

Each task in the data set has two graph versions, grounded and lifted, as explained earlier. For each version, the size of the graphs has a very skewed distribution (whereas the sparsity distribution is relatively flat), with some graphs being particularly large. Table~\ref{tab:data.set} suggests that the lifted version is generally larger than the grounded one. However, because the distribution is rather skewed, we plot additionally in Figure~\ref{fig:size.comparison} the individual graph sizes to offer a complementary view.
The plot indicates that the lifted version is much smaller for the tasks with the largest grounded graphs.


\subsection{Training Details}
For the training of the neural networks, we use the Adam optimizer~\cite{Kingma2015}
with learning rate 0.001. We slightly tune other hyperparameters: the number of layers in GCN and steps in GG-NN is selected from $\{2,4,6\}$ and the dimension of the node representations $h_v^{(t)}$ is chosen from $\{100,150,200\}$. Meanwhile, we also tune the architecture through experimenting with a variant of the attention weights:  replace $w_{\text{gate}}$ in~\eqref{eqn:alpha} by using $D$ separate copies, one for each planner (see the subsection ``Variants'').

We used eight CPU cores and one GPU for training. The consumed memory was approximately 5--10GB. The training of one model with one data split took approximately 10 minutes.

\subsection{Effectiveness of Graph Neural Networks}

\subsubsection{Single planner; Delfi split:} We compare the performance of several types of methods, as summarized in Table~\ref{tab:compare.baseline}. 
%
In addition to the coverage---the percentage of solved tasks---the column ``eval.\ time'' presents the time needed for selecting a planner, which includes, for example, the time to convert a planning task to a graph and that to evaluate the neural network model. This time is overhead and hence for any reasonable method, it should not occupy a substantial portion of the overall time allowance $T=1800$s.


\begin{table}[ht]
\centering
\caption{Percentage of solved tasks in the test set and average evaluation time of the method. Delfi split; single planner.}
\label{tab:compare.baseline}
\begin{tabular}{lcc}
\hline
Method & Solved & Eval. Time \\
\hline
Random planner               & 60.6\% & 0 \\
Single planner for all tasks & 64.8\% & 0 \\
\hline
Complementary2               & 84.8\% & 0\\
Planning-PDBs                & 82.0\% & 0\\
Symbolic-bidirectional       & 80.0\% & 0\\
\hline
Enhanced features + random forest & 82.1\% & 0.51s\\
\hline
Image based, CNN, grounded   & 73.1\% & 11.00s \\
Image based, CNN, lifted     & 86.9\% & 3.16s \\
Graph based, GCN, grounded   & 80.7\% & 23.15s\\
Graph based, GCN, lifted     & \textbf{87.6\%} & 9.41s\\
Graph based, GG-NN, grounded & 77.9\% & 14.53s\\
Graph based, GG-NN, lifted   & 81.4\% & 11.44s\\
\hline
\end{tabular}
\end{table}

The first two methods are weak baselines. As the name suggests, ``random planner'' uniformly randomly selects a planner, whereas ``single planner for all tasks'' uses the one that solves the most number of tasks in the training set. Neither method takes time to perform selection. The percentage of solved tasks for the random method is the expected value.

The next three are state-of-the-art planing systems, not based on deep
learning. These systems are the top performers of IPC 2018, second
only to Delfi. Both Complementary2 and Planning-PDBs perform
$A^{\ast}$ search with heuristic guidances based on sophisticated
methods for pattern databases
creation \cite{franco-et-al-ipc2018b,moises-et-al-ipc2018}.
Symbolic-bidirectional, on the other hand, is a baseline entered into
the competition by the organizers. As the name suggests, it runs a
bidirectional symbolic search, with no heuristic
guidance \cite{torralba-et-al-aij2017}.  None of these methods is
portfolio based and hence no time is needed for planner
selection. Still, they are rather competitive for cost-optimal
planning.

The next one is a machine learning approach based on enhanced features
developed by~\cite{Fawcett2014}. With these handcrafted features, any
standard, non-deep, machine learning model may be applied. We use
random forest, reportedly the best model experimented by the authors
of the referenced paper. In addition to the percentage coverage, time
to compute the features is reported in the table.

Followed are deep learning methods: the two CNNs come from Delfi and the GCNs and GG-NNs are our proposal. For each network architecture, the performance of using grounded/lifted inputs are separately reported.

The results in Table~\ref{tab:compare.baseline} show that the planners in the portfolio have good qualities: with close to twenty planners, even a random choice can solve more than $60\%$ of the tasks. Meanwhile, the state-of-the-art methods, even though not based on deep learning, set a high bar. Delfi, based on CNN, yields a better result for the lifted graphs, but not so much for the grounded ones. Further, one of our GNNs (GCN on lifted graphs) achieves the best performance, whereas the other three GNNs outperform CNN on grounded graphs.

Using either CNNs or GNNs, it appears consistently that the lifted graphs yield better results than do the grounded ones. Moreover, for the same neural network, they also require less evaluation time. One reason is that lifted graphs are less expensive to construct, albeit being larger on average.

We confirm from the table that for all deep learning methods, the time for selecting a planner is negligble compared with the allowed time for executing the planner.

\subsubsection{Single Planner; multiple splits:} We additionally report in Table~\ref{tab:perf.with.std} the results of multiple splits, for the lifted graphs, following~\cite{Sievers2019}. See the top three rows. Similar to the above observations, GCN consistently works better than GG-NN and it also outperforms CNN. Moreover, one generally obtains better performance by using random splits, compared with domain-preserving ones. 

%
%
%

\begin{table}[ht]
\centering
\caption{Percentage of solved tasks in the test set (lifted version). Multiple splits; single planner.}
\label{tab:perf.with.std}
\begin{tabular}{lcccc}
\hline
& \multicolumn{2}{c}{Domain-preserv.} &
\multicolumn{2}{c}{Random}\\
& Mean & Std & Mean & Std\\
\hline
Image based, CNN   & 82.1\% & 6.6\% & 86.1\% & 5.5\%\\
Graph based, GCN   & \textbf{85.6\%} & \textbf{5.5\%} & \textbf{87.2\%} & \textbf{3.5\%}\\
Graph based, GG-NN & 76.6\% & 5.8\% & 74.4\% & 2.7\%\\
\hline
Adaptive, GCN      & \textbf{91.1\%} & \textbf{3.8\%} & \textbf{92.1\%} & \textbf{3.2\%}\\
Adaptive, GG-NN    & 83.0\% & 5.8\% & 86.6\% & 2.0\%\\
\hline
\end{tabular}
\end{table}

\subsection{Effectiveness of Adaptive Scheduling}
\begin{figure}[ht]
\centering
\includegraphics[width=0.9\linewidth]{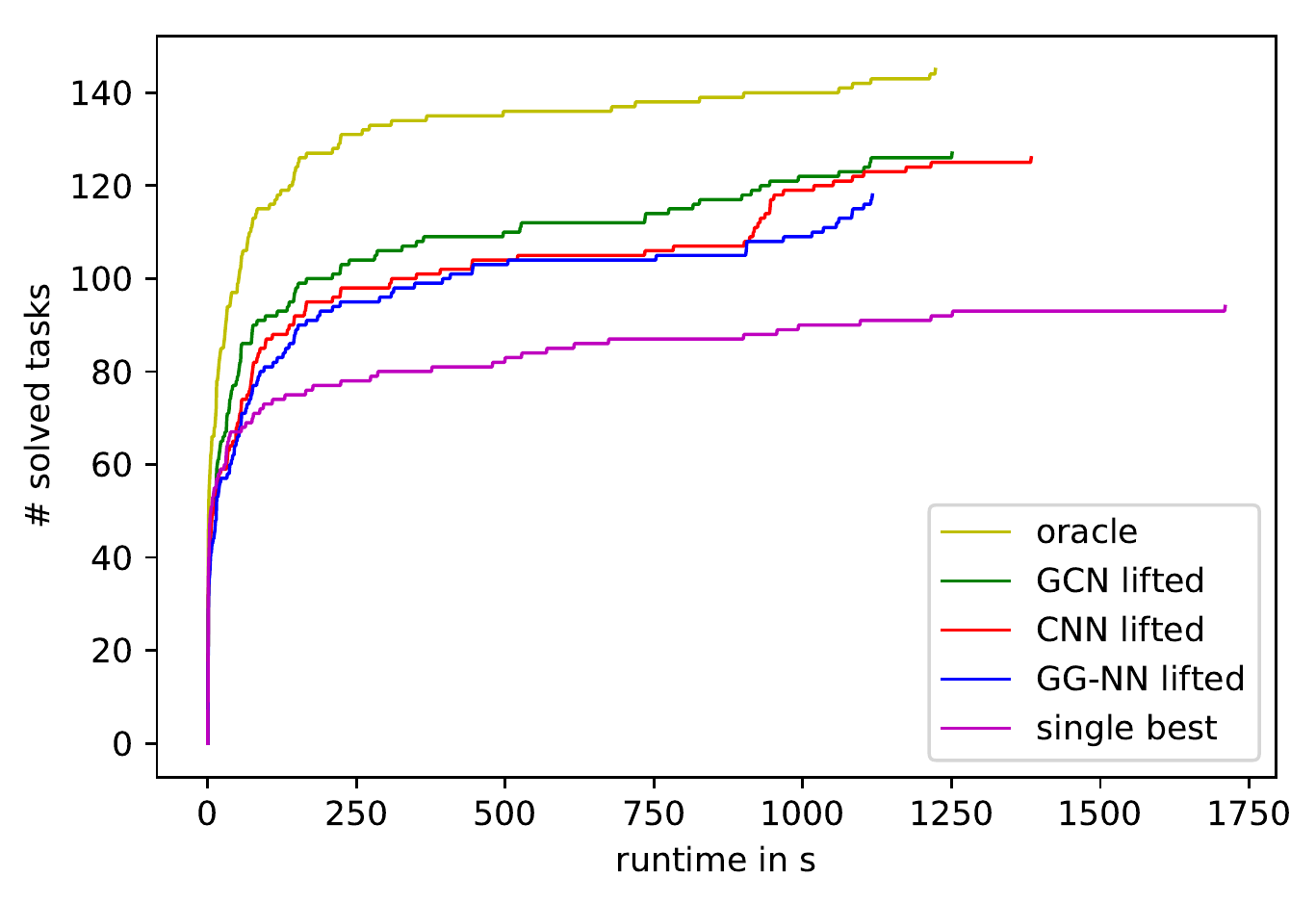}
\caption{Number of solved tasks with respect to time. ``Oracle'' is the ceiling performance obtained by always selecting the fastest planner for each task. Curves for the grounded graphs are not shown to avoid cluttering.}
\label{fig:compare.baseline}
\end{figure}

We first verify the motivation of using more than one planner: frequently a planner solves a task rather quickly, if it ever does. In Figure~\ref{fig:compare.baseline} we plot a curve for each of the deep learning methods regarding the number of solved tasks as time evolves (to avoid cluttering, only those with the lifted version are shown). The \emph{oracle} curve on the top is the ceiling performance obtained by always selecting the fastest planner for each task. For all curves, one sees a trend of diminishing increase, indicating that most of the tasks are solved rather quickly.

\subsubsection{Two planners; Delfi split:} Based on this observation, allowing halftime for a second chance suffices for an alternative planner to complete the task. Hence, we compare the performance of the single selection with that of adaptive scheduling. For a straightforward variant, we also consider a two-planner fixed scheduling, whereby two planners with the smallest failure probability are selected for execution, each given a timeout limit $T/2$.

\begin{figure}[ht]
\centering
\includegraphics[width=0.8\linewidth]{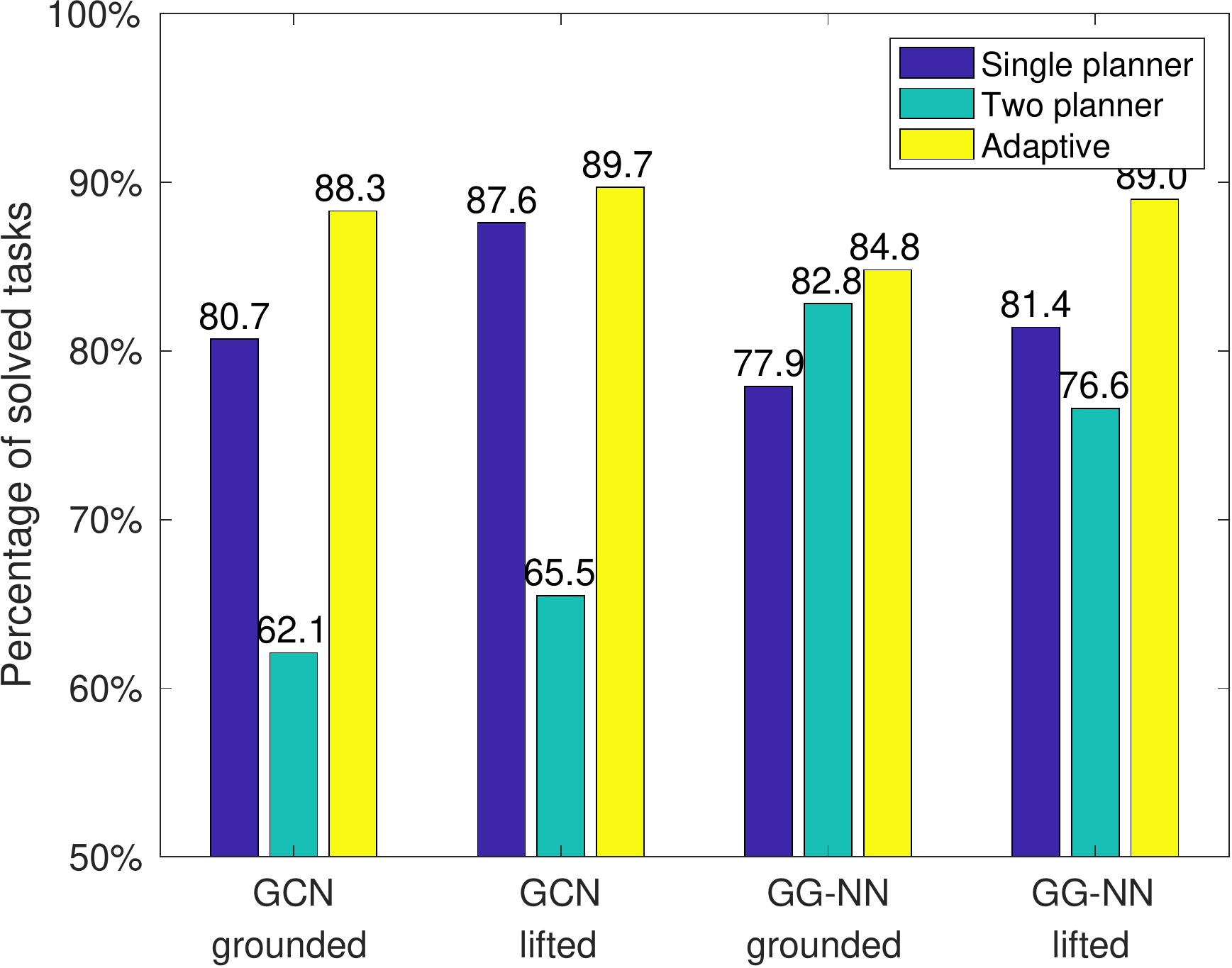}
\caption{Percentage of solved tasks in the test set. Delfi split; two planners.}
\label{fig:adaptive.scheduling}
\end{figure}

Figure~\ref{fig:adaptive.scheduling} reports the results for the Delfi split. One sees that adaptive scheduling consistently increases the percentage of solved tasks on both GNN architectures and both graph versions. It pushes the best performance 87.6\% seen in Table~\ref{tab:compare.baseline} to the new best 89.7\%. On the other hand, two-planner fixed scheduling is generally less competitive than even one-planner. This result is not surprising, because if the single-planner model $f$ is not sufficiently accurate, one does not expect that selecting two may help. The adaptive model $g$, on the other hand, adapts to the case when the first planner fails and hence is more useful.

\subsubsection{Two planners; multiple splits:} For multiple splits, the results are reported in the bottom two rows of Table~\ref{tab:perf.with.std}. One similarly sees that adaptive scheduling consistently improves over single-planner scheduling.

\subsubsection{Multiple planners; Delfi split; simple greedy:} The success of adaptive scheduling raises much interest in the use of more than one planner. A straightforward extension of the adaptive model $g$ is to select more than two planners; but as the number of planners increases, the model becomes more and more complex and challenging to train. Instead, we explore offline, non-machine learning selection methods.

In Table~\ref{tab:offline.portfolio} we consider the following approach: $k$
planners are selected (independent of the task) and each is allotted $T/k$ time.
The first selected planner is the best performer on the training+validation set in the allotted time. The next planners are
chosen in the same manner on the portion of the set not solved by previously selected planners. The top four rows of
Table~\ref{tab:offline.portfolio} report the results. For reference, the next four rows
report the ceiling performance were the planners selected
from the test set.

If $k=2$ planners are used, the coverage is 85.5\%, lower than that
achievable by the proposed adaptive scheduling approach, 89.7\% (see
Figure~\ref{fig:adaptive.scheduling}). If one is willing to use more planners, a better result is achieved with $k=3$. However, the coverage does not increase
monotonically with $k$. Moreover, the performance appears to be much sensitive to the change of $k$. Hence, it is challenging to
find the best $k$.

\begin{table}[ht]
\centering
\caption{Percentage of solved tasks in the test set. Offline method. Delfi split. Compare with performance in Figure~\ref{fig:adaptive.scheduling}.}
\label{tab:offline.portfolio}
\begin{tabular}{lc}
\hline
Method & Solved \\
\hline
(Greedy) Best 2 planners from train set & 85.5\% \\
(Greedy) Best 3 planners from train set & 92.4\% \\
(Greedy) Best 4 planners from train set & 89.7\% \\
(Greedy) Best 5 planners from train set & 87.6\% \\
\hline
(Greedy oracle) Best 2 planners from test set & 93.8\% \\
(Greedy oracle) Best 3 planners from test set & 93.8\% \\
(Greedy oracle) Best 4 planners from test set & 93.1\% \\
(Greedy oracle) Best 5 planners from test set & 92.4\% \\
\hline
Fast Downward Stone Soup & 92.4\% \\
\hline
\end{tabular}
\end{table}


\subsubsection{Multiple planners; multiple splits; FDSS:} A more sophisticated offline planner selection approach is Fast Downward Stone Soup~\cite{helmert-et-al-ipc2011}, wherein the selection of planners and their allotted time are optimized according to a certain score function. Its result is reported in the last row of Table~\ref{tab:offline.portfolio}. This result is achieved with four planners. The coverage coincides with the above greedy approach when taking $k=3$.

\section{Conclusion}
Graphs encode the structural information of a planning task. In this work, we have proposed a graph neural network approach for online planner selection. This approach outperforms Delfi, the winner of the Optimal Track of IPC 2018, which treats a planning task as an image and applies CNNs for selecting candidate planners. Our appealing results are owing to the representation power of GNNs that address the lack of permutation invariance and the negligence of node-labeling information in CNNs. 

We have also proposed an adaptive scheduling approach to compensate the inaccuracy of a single predictive model, through offering a  chance for switching planners at halftime, conditioned on the performance of the previously selected one. Such an adaptive approach consistently increases the number of solved tasks.

Overall, it appears that the lifted graph version is advantageous over the grounded one, because of consistently better performance. However, on average they are larger in size and some are particularly enormous. Moreover, the size distribution is highly skewed in both versions. These factors impose substantial challenges for the batch training of the neural networks. An avenue of future research is to investigate more efficient and scalable training approaches.

We have seen that the use of multiple planners beyond two may improve the performance. Another line of future work is to extend the proposed adaptive scheduling to more than two planners. In this case, the training set construction becomes more and more complex and the model may be challenging to train. Additionally, a principled approach of setting the right number of planners is to be developed.

\section*{Acknowledgments}
Patrick Ferber was funded by DFG grant 389792660 as part of TRR~248 (see https://perspicuous-computing.science).

\fontsize{9.5pt}{10.5pt}\selectfont
\bibliographystyle{aaai}
\bibliography{reference3}

\end{document}